\documentclass[screen,nonacm]{acmart}
\AtBeginDocument{%
  }

\setcopyright{rightsretained}
\copyrightyear{2025}
\acmYear{2025}
\acmDOI{XXXXXXX.XXXXXXX}
\acmConference[Conference acronym 'XX]{Make sure to enter the correct
  conference title from your rights confirmation email}{June 03--05,
  2018}{Woodstock, NY}
\acmISBN{978-1-4503-XXXX-X/2018/06}

\usepackage{wrapfig}
\usepackage{tcolorbox}
\usepackage[font=small,skip=0pt]{caption}

\definecolor{MajorelleBlue}{RGB}{97, 60, 246}

\begin{document}



\title{How Individual Traits and Language Styles Shape Preferences\\In Open-ended User-LLM Interaction: A Preliminary Study}






\author{Rendi Chevi}
\author{Kentaro Inui}
\author{Thamar Solorio}
\author{Alham Fikri Aji}
\affiliation{%
  \institution{MBZUAI}
  \city{Abu Dhabi}
  \country{UAE}
}
\email{{rendi.chevi, kentaro.inui, thamar.solorio, alham.fikri}@mbzuai.ac.ae}

\renewcommand{\shortauthors}{Chevi et al.}





\begin{teaserfigure}
    \centering
    \includegraphics[trim={0cm 1.75cm 0cm 2.5cm},clip,width=1.0\linewidth]{"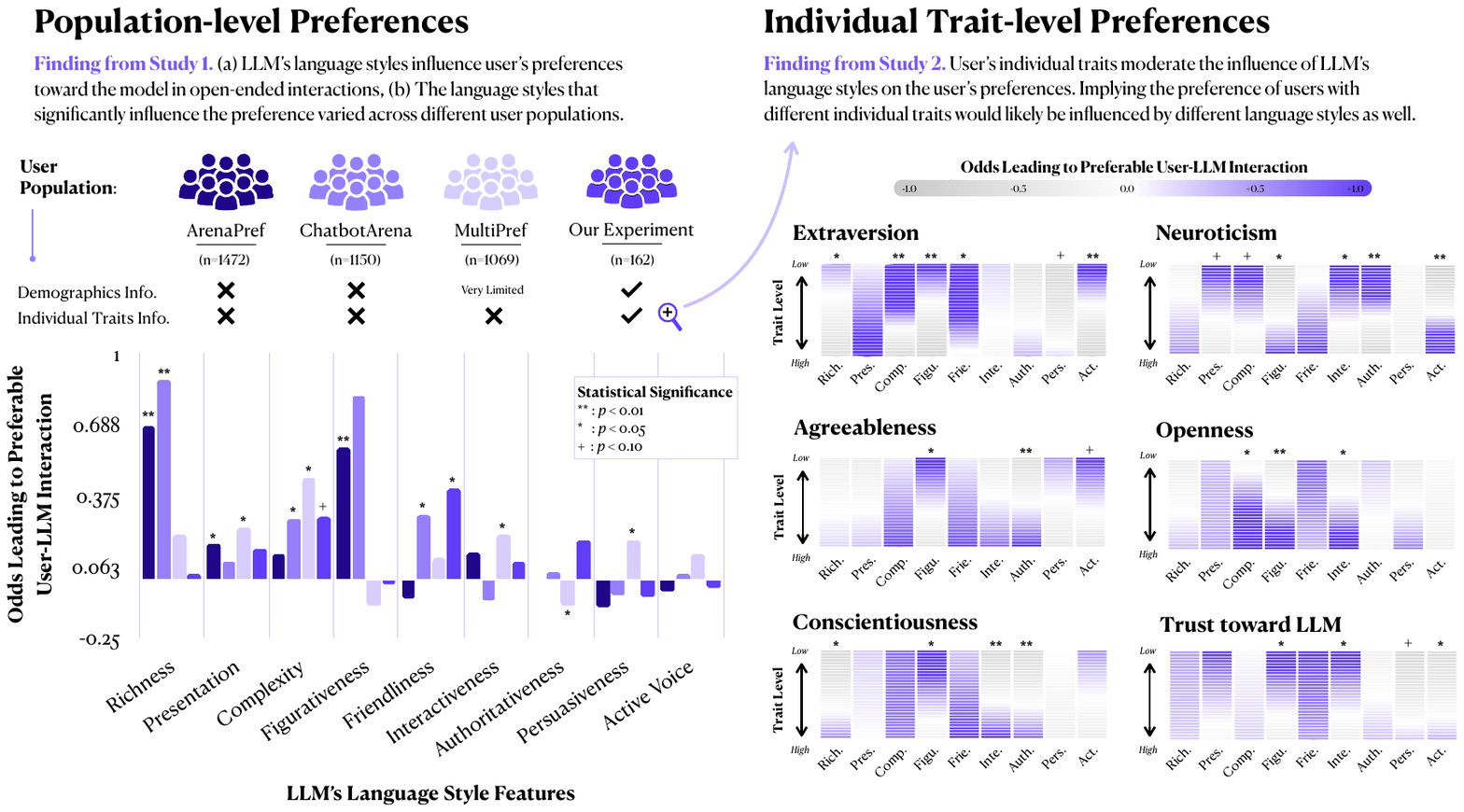"}
    \caption{Our studies explore how the LLM's language styles and user's individual traits influence user's preferences toward the LLM in open-ended interaction. In \textcolor{MajorelleBlue}{\textbf{Study 1}}, we conducted exploratory study on the direct influence of LLM's language styles on user's preferences. In \textcolor{MajorelleBlue}{\textbf{Study 2}}, we conducted experimental study on the moderating effects of user's individual traits on the influence of LLM's language styles on user's preferences.}
    \label{fig:main_results}
\end{teaserfigure}

\maketitle

\section{INTRODUCTION} \label{sec:introduction}

What makes an interaction with the Large Language Model (LLM) more preferable for the user? It is intuitive to assume that information accuracy in the LLM’s responses would be one of the influential variables. While sometimes that is indeed the case~\cite{jung2024we,kabir-stack-overflow-obsolete}, recent studies have found that \textit{inaccurate} LLM's responses could still be preferable when they are not admitting its limitation~\cite{li-etal-2024-dissecting}, perceived to be more authoritative~\cite{metzger_calibrated_distrust}, certain~\cite{kim_uncertainty_llm}, well-articulated~\cite{kabir-stack-overflow-obsolete}, or simply more verbose~\cite{wu-aji-2025-style}. Each of these variables interestingly falls under the category of \textit{language style}~\cite{kang-hovy-2021-style}, as it represents a linguistic feature with a communicative purpose, implying the LLM's language style might be significantly influencing the user's preferences.

Understanding the influence of LLM's language style on user's preferences is crucial, particularly in \textit{open-ended interaction} mode---where users deliberately converse and exchange new information and opinions with the LLM, as it could lead to double-edged consequences. Having the LLM conversing with the style that maximizes user's preferences might be valuable in improving overall user experiences~\cite{volkel2021examining,shumanov2021making}. But, it also means the users would be more susceptible to accepting information from the LLMs that might be misinformed or hallucinated~\cite{bommasani2021opportunities,sahoo-etal-2024-comprehensive}. How does the relationship between the LLM's language style and user's preference really work? Are all users being influenced similarly by the same language styles? Or perhaps, user's personal factors such as their demographics and individual traits also play a role in shaping their preferences? Our long-term objective is to address these research inquiries. As a starting point, we frame these inquiries into the following research questions (RQs):


\begin{itemize}
    \setlength{\itemindent}{-2em}
    \item \textbf{RQ.1}: How do the LLM's language styles influence the user's preferences toward their interaction with the model?
    \item \textbf{RQ.2}: How do the user's individual traits moderate the influence of the LLM's language styles\\on the user's preferences?
\end{itemize}

In the following sections, we will answer our RQs through a series of exploratory and experimental user studies.






\section{STUDY 1: Exploratory Study on Language Style and Preference in User-LLM Interaction} \label{sec:study_1}

To answer \textbf{RQ.1}, we conduct an exploratory study on three \textit{preference-alignment} datasets: ArenaPref~\cite{chiang2024chatbot}, MultiPref~\cite{miranda2024hybrid}, and ChatbotArena~\cite{zheng2023judging}. These secondary datasets contain a wide-variety of real-world user-LLM interactions, each representing different user populations. However, they do not contain user-specific information, such as the user's individual traits. Nonetheless, they have been prevalent in shaping the research of the human-LLM preference alignment~\cite{donyehiya2024futureopenhumanfeedback}.



\noindentparagraph{\textbf{User-LLM Interaction Data Selection.}} \label{sec:study_1_data_selection}

Each instance in the dataset is composed of a user's query, two different LLM's responses, and the user's binary preference for the responses. We focused only on open-ended interaction scenarios and found that including only queries with interrogative prefixes (e.g. \textit{what}, \textit{how}, \textit{are}) and without math, code, or computation keywords, effectively filters out non-open-ended scenarios. To minimize confounding variables, we further constrain our instances to only those that are in English, involve single-turn interactions, and have response pairs that are semantically similar by measuring their text embeddings' cosine similarity.




\noindentparagraph{\textbf{Stylistic Features Measurement.}} \label{sec:study_1_measurement}

Drawing from works in the linguistic analysis of the LLMs and language style in general~\cite{li-etal-2024-dissecting,li2024generative,danescu-niculescu-mizil-etal-2013-computational}, we define nine style features to be measured: information \textit{richness}, information \textit{presentation}, vocabulary \textit{complexity}, usage of \textit{active voices}, \textit{figurativeness}, \textit{friendliness}, \textit{interactiveness}, \textit{authoritativeness}, and \textit{persuasiveness}. We implemented a Natural Language Processing (NLP) pipeline to measure the intensity level of each style feature. Depending on the implicitness of the style, we either measure them via rule-based NLP algorithms or neural-based models (details in Appendix~\ref{sec:app_measurement_style}). 


\noindentparagraph{\textbf{Binary Preference Regression Analysis.}} \label{sec:pref_regression}

To analyze the influence of the measured LLM's style features on user's preferences, we fitted a binary preference regression analysis~\cite{peysakhovich2015pairwise} on each user population. Let $x_a,x_b \in \mathbb{R}^m$ denote the style feature pairs of response $a$ and $b$, where $m=9$ is the number of style features, and $y$ be the preference toward $a$ or $b$. We defined the independent variables as the difference between the style features, $x = x_a - x_b$, and the dependent variables as $y \in \{0,1\}$, where $y=1$ if the user prefers response $a$, and $y=0$ otherwise. Our preference regression model is then defined as: $y = logit(\beta_0 + \sum_{i=1}^9 \beta_ix_i)$.

\subsection{Exploratory Study Findings}

We examined the parameters of the fitted preference regression models, particularly the odds associated with the statistically significant ($p < 0.05$) style feature's coefficient, $1 - \exp(\beta_i)$, which describes the change of the user's preference resulting from an increase in a style feature's intensity. We visualize the result in Fig~\ref{fig:main_results}, detailed result is attached in Appendix~\ref{sec:app_regression_result}.


\noindentparagraph{\textbf{LLM's Language Style Does Influence User's Preference.}}

We found that there are at least 3 statistically significant style features influencing user's preferences across user populations. In ArenaPref population, an increase in \textit{Richness} ($\scriptstyle \uparrow$88.6\%), \textit{Complexity} ($\scriptstyle \uparrow$26.9\%), and \textit{Friendliness} ($\scriptstyle \uparrow$28.9\%) in the LLM's responses elevate user's preference. In ChatbotArena, it was \textit{Richness} ($\scriptstyle \uparrow$68.3\%), \textit{Presentation} ($\scriptstyle \uparrow$16.0\%), and \textit{Figurativeness} ($\scriptstyle \uparrow$58.1\%). Meanwhile, MultiPref seemed to be influenced by a more diverse set of styles, where an increase in \textit{Presentation} ($\scriptstyle \uparrow$23.7\%), \textit{Complexity} ($\scriptstyle \uparrow$45.9\%), \textit{Interactiveness} ($\scriptstyle \uparrow$20.0\%), \textit{Persuasiveness} ($\scriptstyle \uparrow$17.9\%), along with a decrease in \textit{Authoritativeness} ($\scriptstyle \downarrow$11.1\%) would likely elevate user's preference. 

\noindentparagraph{\textbf{Population-level Preference Influenced by Different LLM's Language Style.}}

While the LLM's language style does have influence on user's preference, it varies across different user populations. This variation implies the presence of confounding variables outside of ones we've controlled in~§\hyperref[sec:study_1_data_selection]{2a}, and possible moderating variables such as the user's demographics and traits that are unknown in the datasets, which we are about to investigate in the next study.


\begin{figure}[!t]
    \centering
    \includegraphics[trim={3cm 13.5cm 2.8cm 2.25cm},clip,width=1.0\linewidth]{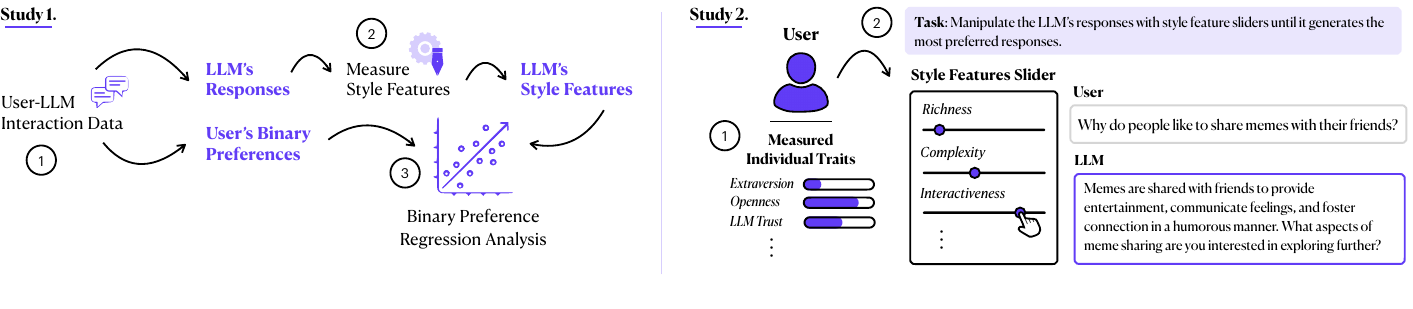}
    \caption{Methodological overview of \textcolor{MajorelleBlue}{\textbf{Study 1}} (Left) and \textcolor{MajorelleBlue}{\textbf{Study 2}} (Right). Experimental GUI for Study 2 is attached in Appendix~\ref{sec:app_gui}.}
    \label{fig:method}
\end{figure}

\section{STUDY 2: Experimental Study on Individual Traits, Style, and Preference in User-LLM Interaction}

To answer \textbf{RQ.2}, we conducted an experimental study involving human users. We deliberately collect and study the role of user's individual traits, a variable that tends to be missing in the current User-LLM preference datasets and study.  


\noindentparagraph{\textbf{Experimental Design and Procedure.}}

We designed a within-subject experiment where each user went through 2 stages. In the first stage, we asked them to fill-in questionnaires that measure their individual traits. In the second stage, we asked them to interact with an LLM via the provided interface, they will be given the ability to manipulate the LLM's language styles, and asked to manipulate the LLM's responses to their preferences. The core methods in this experiment follows \textit{sampling with people} methodology~\cite{markov_chain_with_people,harrison2020gibbs}, which we designed to \textit{sample} the LLM's language style that maximizes the user's preferences. 


\noindentparagraph{\textbf{User Participants.}}


Our user participants pool are based in the UK, use English as their primary language, within the ages of 20-30, balanced by sex, and are daily users of LLM services (e.g. OpenAI, Anthropic). We recruited 10 users from the Prolific platform, where each user contributed 60 samples of preferences. After filtering, we have a total of 162 valid preference samples to analyze. Detailed user's statistics are reported in Appendix~\ref{sec:app_demographics}. Given that we have not covered wider demographic diversity and larger sample sizes, it is important to interpret the findings in this study cautiously.

\noindentparagraph{\textbf{User's Individual Traits Measurement.}}

We measure two sets of user's individual traits: their personality traits and trust toward the LLMs. To measure their personality traits, we administer the 10-item measure of the Big-5 personality dimensions~\cite{gosling2003very} to the user, which measure the user's level of \textit{Extraversion}, \textit{Conscientiousness}, \textit{Neuroticism}, \textit{Openness}, and \textit{Agreeableness}. To measure their trust toward the LLMs, we administer the ChatGPT Trust Scale~\cite{choudhury2023investigating} to the user, where we slightly adapted the scale to refer to LLMs in general, rather than specifically focusing on ChatGPT.

\noindentparagraph{\textbf{Stimuli Design, Style-varying LLM's Responses.}}

We first defined 3 queries for this study and prompt a LLM (OpenAI’s GPT-4o-Mini) with a factual context to provide a baseline response. To craft responses in a variety of styles, we implemented a zero-shot style transfer pipeline following~\cite{reif-etal-2022-recipe}, which we designed to modify the baseline response to convey a given style feature. As we have 3 queries and 9 style features with 3 intensity levels, we synthesized a total of $3 * 3^{9} = 59,049$ LLM’s responses as the possible stimuli for the users. Details of the style transfer prompts is attached in Appendix~\ref{sec:app_prompt}. 

\noindentparagraph{\textbf{Sampling Preference-eliciting Style with People.}}

Collecting user's preferences in a similar fashion as the datasets used in Study 1 (§\ref{sec:study_1}) would be prohibitively costly for us to do. We instead adopted \textit{gibbs sampling with people}~\cite{harrison2020gibbs} methodology to effectively sample the LLM's responses with style features that maximize each user's preferences. Let $g(v_1,...,v_9)$ be the LLM's response parametrized by the style features, we ask the user to iteratively manipulate the intensity of $v_i \in \{1,2,3\}$ while other feature intensities are fixed, then choose which intensity generates LLM's responses that they prefer the most. In the end, we would have chains of style features that converged toward a style combination the user prefers the most.




\noindentparagraph{\textbf{Moderated Binary Preference Regression Analysis.}}

To analyze the moderation effect of user's individual traits, we expand the regression analysis in Study 1 to include the traits as moderator variables. For each measured individual trait, $z_k$, we fitted a moderated preference regression model defined as: $y = logit(\beta_0 + \sum_{i=1}^9 \beta_ix_i + \sum_{j=1}^9 \beta_jx_iz_k)$.

\subsection{Experimental Study Findings}

For each trait $z_k$, We examined the shift of odds associated with the statistically significant ($p < 0.05$) trait-moderated style feature's coefficient, $1 - exp(\beta_i + \beta_jz_k)$. We visualize the result in Fig.~\ref{fig:main_results} (Right).




\noindentparagraph{\textbf{Individual Traits Moderate the Influence of Language Style Differently.}}

\begin{wrapfigure}{R}{0.25\textwidth}
\centering
\includegraphics[trim={0 10cm 23.5cm 6.25cm},clip,width=0.25\textwidth]{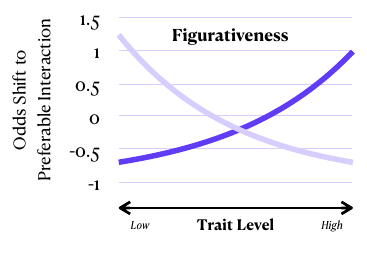}
\captionsetup{justification=raggedleft}
\caption{\textbf{Example of Polarizing Moderator Effects}}
\label{fig:diff_effects_traits}
\end{wrapfigure}

We found that depending on user's individual traits, certain LLM's language styles influence user's preferences differently. For users with higher level of $\scriptstyle \uparrow$\textit{Agreeableness}, lower level of $\scriptstyle \downarrow$\textit{Figurativeness} and higher level of $\scriptstyle \uparrow$\textit{Authoritativeness} increase their preferences. For users with $\scriptstyle \uparrow$\textit{Extraversion}, it was $\scriptstyle \downarrow$\textit{Figurativeness}, $\scriptstyle \downarrow$\textit{Richness}, $\scriptstyle \downarrow$\textit{Complexity}, $\scriptstyle \downarrow$\textit{Friendliness}, and $\scriptstyle \downarrow$\textit{Active Voice}. Whereas for users with $\scriptstyle \uparrow$\textit{Neuroticism}, it was $\scriptstyle \uparrow$\textit{Figurativeness}, $\scriptstyle \uparrow$\textit{Active Voice}, $\scriptstyle \downarrow$\textit{Authoritativeness} and $\scriptstyle \downarrow$\textit{Interactiveness}. For users with $\scriptstyle \uparrow$\textit{Openness}, it was $\scriptstyle \uparrow$\textit{Figurativeness}, $\scriptstyle \uparrow$\textit{Complexity}, and $\scriptstyle \uparrow$\textit{Authoritativeness}. Meanwhile, users with $\scriptstyle \uparrow$\textit{Trust toward LLMs} are influenced by $\scriptstyle \downarrow$\textit{Figurativeness}, $\scriptstyle \downarrow$\textit{Interactiveness}, and $\scriptstyle \uparrow$\textit{Active Voice}.

\noindentparagraph{\textbf{Polarizing Effects of Individual Traits.}}

Though we have observed various statistically significant moderation effects of each trait independently, user's individuality is represented by a combination of these traits as whole. In the case of \textit{Extraversion} and \textit{Neuroticism} for example, we can see that these traits moderate user's preference in a polarizing way (Fig.~\ref{fig:diff_effects_traits}). How do these dynamics apply to users with both high and low levels of those traits? Which language styles will influence them more? Future studies could explore these questions further with more observational samples and applying techniques such as joint moderation effects or other explainability methods~\cite{shap}.




\section{Conclusion, Limitation, and Future Direction}

In this paper, we presented our preliminary study on how user's very own individual traits and LLM's language style influence user's preferences in open-ended User-LLM interaction. As a preliminary study, it is important to interpret our findings with caution, given that our samples still need wider demographic diversity and larger sample sizes. Our future direction is to first address these limitations. We are then interested to further investigate the joint effects and possible causal relationship between and \textit{beyond} our variables; Why do users with certain traits are more or less influenced by certain language style? Will the elicited preferences actually translate into positive outcomes for the users? Or will it exacerbate other pervasive outcomes such as user's susceptibility to misinformation~\cite{zhou-synthetic-lies,pat-slip-through,govers-feeds-distrust} and negative psychosocial effects~\cite{liu2024chatbot, fang2025ai, phang2025investigating}.

\bibliographystyle{ACM-Reference-Format}
\bibliography{biblio}

\appendix

\section{Details on Study 1} \label{sec:study_1_details}

\subsection{Measurement of Language Style Features} \label{sec:app_measurement_style}

In Table~\ref{tab:style_measurement_details}, we listed the measurement methods for each language style features we have defined. This collection of methods make up our style feature measurement pipeline mentioned in~§\ref{sec:study_1_measurement}. In Table~\ref{tab:style_measurement_summary}, we reported the summary statistics (mean and standard deviation) of the measured stylistic features in the LLM's responses across populations. 

\vspace{0.33cm}

\begin{table}[!htb]
    \footnotesize
    \centering
    \begin{tabular}{p{0.15\linewidth}|p{0.45\linewidth}|p{0.15\linewidth}}
    \toprule
         \textbf{Style}& \textbf{Measurement Method} &\textbf{Main Tools}\\
    \midrule
         Richness& Measure the frequencies of the nouns, adjectives, conjunctions, coordinating conjunctions, and subordinating conjunctions in the LLM's response.&spaCy's NLP pipeline\\
         \midrule
         Presentation& Measure the presence of various markdown styling, such as bolding, italicizing, and list enumeration formatting in the LLM's response.&RegEx matching\\
 \midrule
 Complexity&Measure the Dale-Chall readability score of the LLM's response.&Textstat library\\
 \midrule
 Figurativeness&Measure the intensity level of figurativeness through zero-shot classification prompting.&OpenAI's GPT-4o-Mini as zero-shot classifier\\
 \midrule
 Friendliness&Measure the intensity level of friendliness through zero-shot classification prompting.&OpenAI's GPT-4o-Mini as zero-shot classifier\\
 \midrule
 Interactiveness&Measure the intensity level of interactiveness through zero-shot classification prompting.&OpenAI's GPT-4o-Mini as zero-shot classifier\\
 \midrule
 Authoritativeness&Measure the intensity level of authorativeness through neural-based classifier model.&BERT-based model trained on Szeged Uncertainty Corpus\\
 \midrule
 Persuasiveness&Measure the discrete intensity level of persuasiveness through zero-shot classification prompting.&OpenAI's GPT-4o-Mini as zero-shot classifier\\
 \midrule
 Active Voice&Measure the frequencies of the linguistic pattern match of active and passive voices in the LLM's response.&spaCy's NLP pipeline and linguistic pattern matching\\
 \midrule
    \end{tabular}
    \caption{Measurement methods and tools for each language style features.}
    \label{tab:style_measurement_details}
\end{table}



\begin{table}[!htb]
    \footnotesize
    \centering
    \begin{tabular}{llllllllll}
    \toprule
           &\multicolumn{9}{c}{Mean (SD) of Stylistic Features}\\
    \midrule
  & \textbf{Rich.}& \textbf{Pres.}& \textbf{Comp.}& \textbf{Figu.}& \textbf{Frie.}& \textbf{Inte.}& \textbf{Auth.}& \textbf{Pers.}&\textbf{Acti.}\\
  & [0, $\infty$]& [1.0,3.0]& [4.9, 9.9]& [1.0,3.0]& [1.0,3.0]& [1.0,3.0]& [0.0,1.0]& [1.0,3.0]&[0.0,1.0]\\
          \midrule
           ArenaPref&45.83 (37.75)&  0.35 (0.50)&  9.28 (4.69)&  1.01 (0.14)&  1.12 (0.39)&  1.17 (0.47)&  0.64 (0.38)&  1.12 (0.33)& 0.79 (0.29)\\
           ChatbotArena&42.10 (36.07)&  0.28 (0.45)&  8.88 (2.10)&  1.01 (0.11)&  1.07 (0.29)&  1.10 (0.37)&  0.66 (0.38)&  1.13 (0.33)& 0.77 (0.31)\\
  MultiPref& 49.14 (34.54)& 0.35 (0.49)& 9.00 (1.24)& 1.00 (0.09)& 1.18 (0.41)& 1.15 (0.38)& 0.49 (0.37)& 1.46 (0.53)&0.84 (0.23)\\
 \midrule
    \end{tabular}
    \caption{Summary statistics of the measured language style features.}
    \label{tab:style_measurement_summary}
\end{table}

\newpage
\subsection{Binary Preference Regression Results} \label{sec:app_regression_result}

In Table~\ref{tab:study_1_odds}, we attached the more detailed numerical results version of Fig.~\ref{fig:main_results} (Left).
\vspace{0.33cm}

\begin{table}[!htb]
    \footnotesize
    \centering
    \begin{tabular}{llllllllll}
    \toprule
          &\textbf{Rich.}&  \textbf{Pres.}&  \textbf{Comp.}&  \textbf{Figu.}&  \textbf{Frie.}&  \textbf{Inte.}&  \textbf{Auth.}&  \textbf{Pers.}& \textbf{Acti.}\\
 \textbf{Population}& & & & & & & & &\\
          \midrule
          ArenaPref&$0.680^{**}$&  $0.160^{*}$&  0.117&  $0.581^{**}$&  -0.080&  0.125&  -0.004&  -0.121& -0.050\\
          ChatbotArena&$0.880^{**}$&  0.085&  $0.269^{*}$&  0.810&  $0.289^{*}$&  -0.089&  0.034&  -0.062& 0.025\\
 MultiPref& 0.199& $0.230^{*}$& $0.450^{*}$& -0.110& 0.100& $0.200^{*}$& $-0.110^{*}$& $0.179^{*}$&0.116\\
 \midrule
 Our Experiment (Study 2)& 0.031& 0.140& $0.278^{+}$& -0.017& $0.402^{*}$& 0.081& 0.178& -0.070&-0.034\\
 \midrule
    \end{tabular}
    \caption{Odds of each language style features in influencing user's preferences. $^{**}: p < 0.01, ^{*}: p < 0.05, ^{+}: p < 0.1$.}
    \label{tab:study_1_odds}
\end{table}

\section{Details on Study 2} \label{sec:study_2_details}

\subsection{Sampling with People Interface} \label{sec:app_gui}

\begin{figure}[!htb]
    \centering
    \includegraphics[trim={5cm 4cm 5cm 4cm},clip,width=0.80\linewidth]{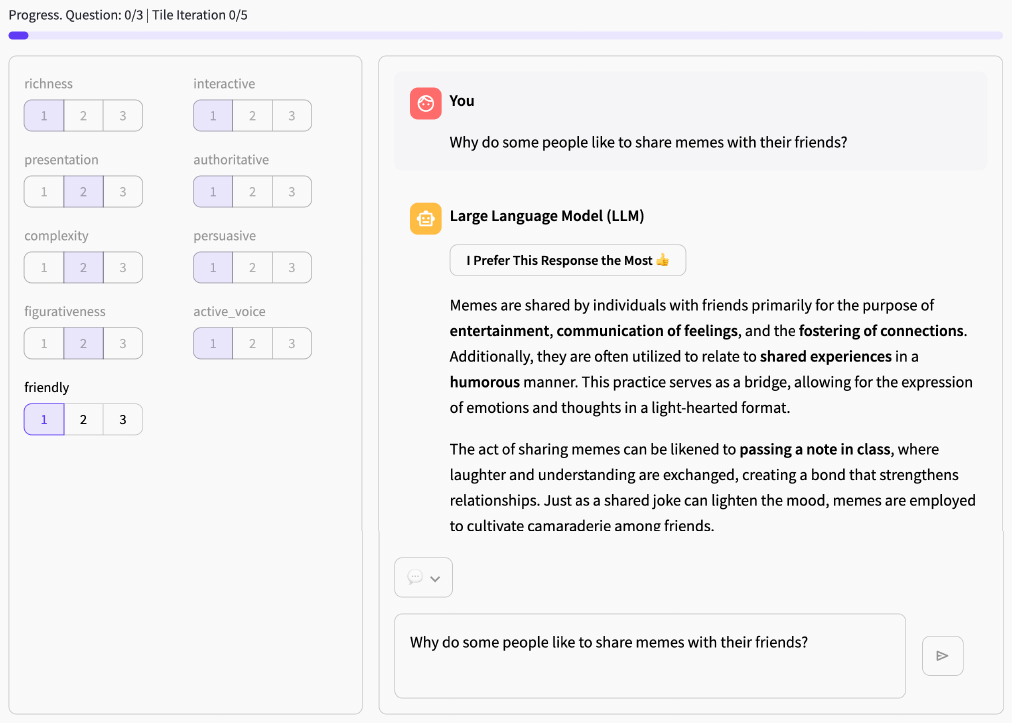}
    \caption{User interface for our sampling with people experiment. During the actual experiment, the style names and intensity levels are not shown to the users, and the order of both are randomized for every iteration.}
    \label{fig:swp_interface}
\end{figure}

The user interface for sampling with people experiment is shown in Fig.~\ref{fig:swp_interface}. Each participant is given the following instruction to follow (the instruction is self-contained in the experimental GUI, we show it here for brevity):

\begin{tcolorbox}[colback=gray!5!white,colframe=MajorelleBlue!75!black]
\small
Go through every option in Tile \#k, until the LLM gives you the response the you prefer the most among the options. After it gives the response that you feel you prefer the most, click '\textbf{I Prefer This Response the Most.}'

\begin{itemize}
    \item Always re-read the new manipulated response from start to finish.
    \item Ask yourself: "Do I like this new response more than the previous one?"
    \item You don't have to be overly objective or over-analyze your decision in preferring or liking the LLM's responses. In fact, you are encouraged to go with "what feels right".
\end{itemize}
\end{tcolorbox}

\newpage
\subsection{User Participants Demographics} \label{sec:app_demographics}

In Table~\ref{tab:study_2_demographics_dist}, we report our participant pool's demographics and individual traits in Study 2.

\vspace{0.33cm}
\begin{table}[!htb]
    \footnotesize
    \centering
    \begin{tabular}{p{0.2\linewidth}|p{0.2\linewidth}p{0.15\linewidth}}
\toprule
 \multicolumn{3}{c}{\textbf{Participant's Demographics}}\\
 \midrule
         \textbf{Age}&  Mean (SD)& 25.30 (2.83)\\
         &  Range& 20-30\\
         \midrule
         \textbf{Sex}&  Female& 5\\
         &  Male& 5\\
         \midrule
         \textbf{Ethinicity}&  Asian& 1\\
         &  Black& 6\\
         &  White& 1\\
         &  Mixed& 1\\
         &  Prefer not to say& 1\\
         \midrule
 \textbf{Daily Usage of AI/LLM}& Every day&3\\
 & Multiple times every day&7\\
 \midrule
 \textbf{LLM Service Usage}& OpenAI’s ChatGPT&10\\
 (A user can report multiple LLM services they used)& Anthropic’s Claude&5\\
 & Google’s Gemini&4\\
 & Others&8\\
 \toprule
 \multicolumn{3}{c}{\textbf{Participant's Individual Traits}}\\
 \midrule
 Extraversion& Mean (SD)&2.85 (0.92)\\
 & Range&1-5\\
 Agreeableness& Mean (SD)&4.00 (0.77)\\
 & Range&1-5\\
 Conscientiousness& Mean (SD)&4.25 (0.71)\\
 & Range&1-5\\
 Neuroticism& Mean (SD)&2.40 (0.86)\\
 & Range&1-5\\
 Openness& Mean (SD)&3.60 (0.73)\\
 & Range&1-5\\
 Trust toward LLM& Mean (SD)&3.59 (0.37)\\
 & Range&1-4\\
 \toprule
 Num. of Participants& &10\\
 Num. of Rejected Participants& &1\\
 \midrule
 Num. of Preference Samples per Participants& &60\\
 Num. of Gibbs Sampling's Burn-in Period& &2\\
 \midrule
 Final Num. of Preference Samples After Filtering& &\textbf{\underline{n = 162}}\\
 & &\\
 \midrule
    \end{tabular}
    \caption{Statistics of user participant's demographics and individual traits in Study 2.}
    \label{tab:study_2_demographics_dist}
\end{table}

\newpage
\subsection{Zero-shot Style Transfer Prompting} \label{sec:app_prompt}

In Table~\ref{tab:study_2_prompts}, we listed the description of the intensity level for each style features, the descriptions are used to prompt the zero-shot style transfer pipeline for synthesizing our stimuli.

\vspace{0.33cm}
\begin{table}[!htb]
    \footnotesize
    \centering
    \begin{tabular}{p{0.15\linewidth}|p{0.8\linewidth}}
    \toprule
         \textbf{Style}& \textbf{Intensity Level \& Description}\\
    \midrule
         Richness& \textbf{L1.} Provides a straightforward, unembellished answer to the question, focusing solely on the essential information.\\
         & \textbf{L2.} Offers the answer along with additional information that adds context or important details but remains relevant to the question.\\
         & \textbf{L3.} Provides the answer along with excessive details, tangents, or background information that, while interesting, does not directly support the original question.\\
         \midrule
         Presentation& \textbf{L1.} Using a single paragraph to structure the utterance without any formatting elements.\\
 &\textbf{L2.} Using two paragprahs to structure the utterance. Important words and phrases in the utterance are formatted with bold or italic style.\\
 &\textbf{L3.} Using more than two paragprahs to structure the utterance with bolding, italicizing, headings, bullet points, numbered list the key words and phrases as the formatting elements.\\
 \midrule
 Complexity&\textbf{L1.} Using vocabulary of verbs, nouns, adjectives, and adverbs that are very easy to read. Easily understood by an average twelve year old student.\\
 &\textbf{L2.} Using vocabulary of verbs, nouns, adjectives, and adverbs that are moderately difficult to read. Best understood by an average high-school student.\\
 &\textbf{L3.} Using vocabulary of verbs, nouns, adjectives, and adverbs that are very difficult to read. Best understood by university graduates and experienced scholars.\\
 \midrule
 Figurativeness&\textbf{L1.} Does not convey any figurative language.\\
 &\textbf{L2.} Re-emphasize an explanation by figurative language in the form of simple metaphor that introduce direct comparisons using common ideas.\\
 &\textbf{L3.} Re-emphasize an explanation by  figurative language in the form of complex metaphors that introduce imaginative yet relatable ideas.\\
 \midrule
 Friendliness&\textbf{L1.} Does not convey expressions of friendliness.\\
 &\textbf{L2.} Use expression that convey politeness, warmth, approachable, and come off as formal.\\
 &\textbf{L3.} Use expression that convey politeness, warmth, approachable, and come off as informal.\\
 \midrule
 Interactiveness&\textbf{L1.} Does not seek further engagement or clarification regarding the query of the utterance.\\
 &\textbf{L2.} Attempt to engage with the user's curiosity. These include prompting the user to consider a broader context or related topics. \\
 &\textbf{L3.} Attempt to engage with the user's curiosity and intention. These include explicitly asking for more relevant information or seeking to understand the user's intent. \\
 \midrule
 Authoritativeness&\textbf{L1.} Using expressions that are lacking in confidence and detail. These includes the incorporation of tentative languages (e.g., "maybe," "might be," "I think").\\
 &\textbf{L2.}  Does not convey expressions of authoritativeness.\\
 &\textbf{L3.} Using expressions that exudes confidence and expertise. These includes the incorporation of assertive language (e.g., “is,” “will,” “must”).\\
 \midrule
 Persuasiveness&\textbf{L1.} Does not use expressions that attempts to convince the user more to accept the information, statements, facts, or opinions in the utterance.\\
 &\textbf{L2.} Attempts to convince the user more to accept the information or opinions in the utterance, using moderate emotional appeal or reasoning which lacks deeper engagement or urgency.\\
 &\textbf{L3.} Attempts to convince the user more to accept the information or opinions in the utterance, using strong emotional appeal or reasoning to effectively convince the user.\\
 \midrule
 Active Voice&\textbf{L1.} Always using passive voice, if the context is appropriate, aiming for a less direct and less engaging style of communication.\\
 &\textbf{L2.} Using a mix of both passive and active voice, striking a balance between engagement and formality style of communication.\\
 &\textbf{L3.} Always using active voice, if the context is appropriate, resulting in clear, direct, and engaging style of communication.\\
 \midrule

    \end{tabular}
    \caption{Description of language style features' intensity level used for our zero-shot style transfer pipeline to synthesize style-varying LLM's responses.}
    \label{tab:study_2_prompts}
\end{table}

\end{document}